\documentclass{article}
\makeatletter
\let\NAT@parse\undefined
\makeatother
\usepackage[numbers]{natbib}

\usepackage{float}

\usepackage[dvipsnames]{xcolor}

\usepackage[preprint]{neurips_2024}

\usepackage[utf8]{inputenc} %
\usepackage[T1]{fontenc}    %
\usepackage[hidelinks,colorlinks=true,urlcolor=Magenta]{hyperref}       %
\usepackage{url}            %
\usepackage{booktabs}       %
\usepackage{amsfonts}       %
\usepackage{nicefrac}       %
\usepackage{microtype}      %
\usepackage{xcolor}         %
\usepackage{amsmath,amssymb}
\usepackage{mathtools}
\usepackage{graphicx}       %

\newcommand\extrafootertext[1]{%
    \bgroup
    \renewcommand\thefootnote{\fnsymbol{footnote}}%
    \renewcommand\thempfootnote{\fnsymbol{mpfootnote}}%
    \footnotetext[0]{#1}%
    \egroup
}

\usepackage[misc,geometry]{ifsym}

\title{InfiniMotion: Mamba Boosts Memory in Transformer for Arbitrary Long Motion Generation}

\author{
Zeyu Zhang$^{12*}$,~ Akide Liu$^{1}$,~ Qi Chen$^{3}$,~ Feng Chen$^{13}$,~ Ian Reid$^{34}$,\\
\textbf{Richard Hartley$^{2}$,}~ \textbf{Bohan Zhuang$^{1}$,}~ \textbf{Hao Tang$^{5\text{\Letter}}$}
\\ [0.285cm]
$^1$Monash University~ $^2$The Australian National University~ $^3$The University of Adelaide\\
$^4$Mohamed bin Zayed University of Artificial Intelligence~ $^5$Peking University
}

\begin{document}

\extrafootertext{\fontsize{8}{4}\selectfont $^{*}$Work done while being a research assistant at Monash University.}
\extrafootertext{\fontsize{8}{4}\selectfont $^{\text{\Letter}}$Corresponding author: \href{bjdxtanghao@gmail.com}{bjdxtanghao@gmail.com}}

\maketitle

\begin{center}
\vspace{-1cm}
   \url{https://steve-zeyu-zhang.github.io/InfiniMotion} 
\end{center}

\begin{figure}[H]
    \centering
    \href{https://youtu.be/VpVvH26Dza4}{\includegraphics[width=0.9\linewidth]{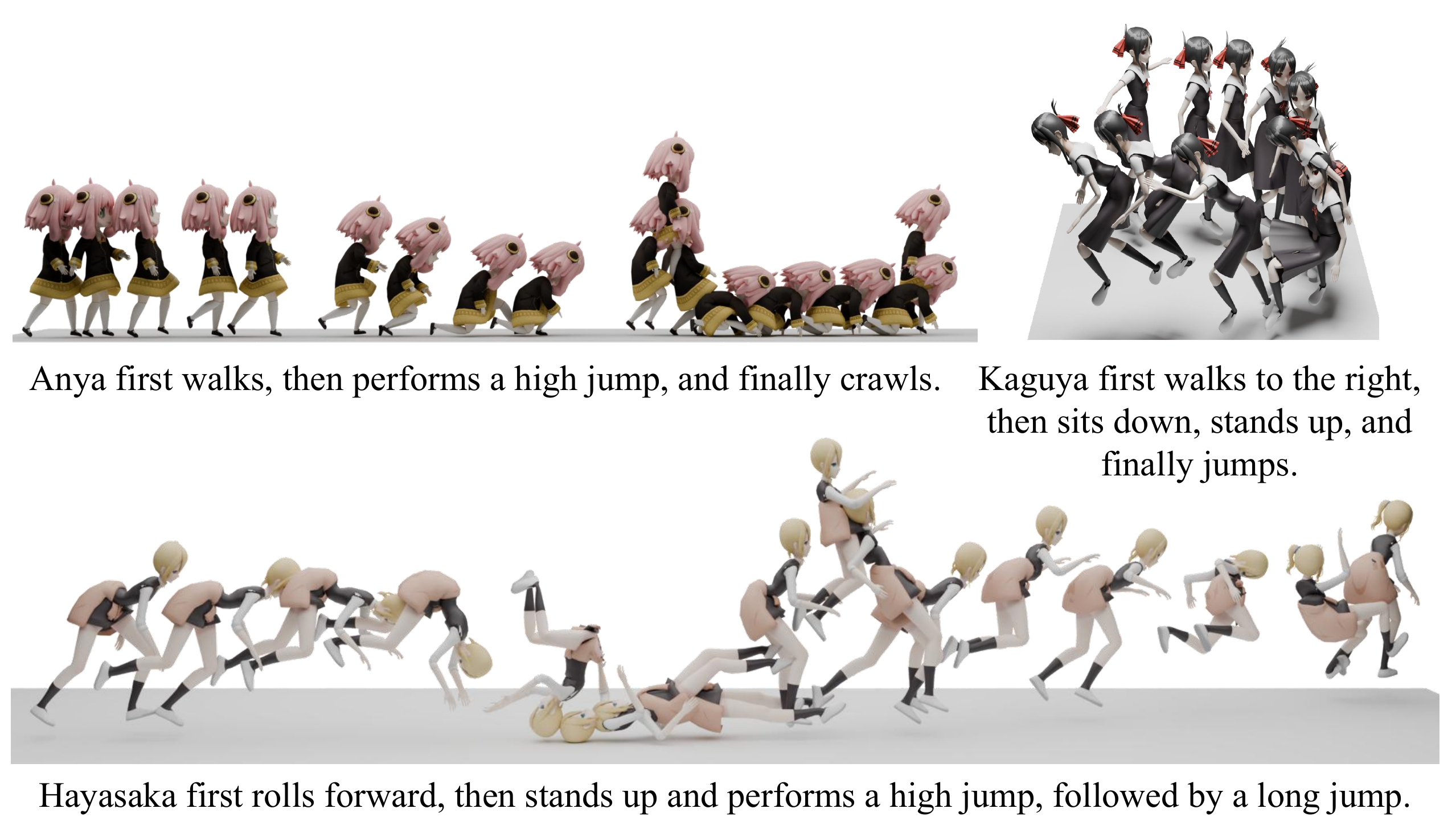}}
    \caption{The diagram illustrates a variety of representative examples of long motion sequences created by our innovative method, \textbf{InfiniMotion}. Each instance within the diagram is based on at least three consecutive user queries, each describing a distinct action. These examples highlight our method's ability to produce high-quality motion sequences, characterized by smooth transitions between different actions. Click on the diagram for a 1-hour demo video.
    }
    \label{fig:main_demo}
\end{figure}

\begin{abstract}

Text-to-motion generation holds potential for film, gaming, and robotics, yet current methods often prioritize short motion generation, making it challenging to produce long motion sequences effectively: (1) Current methods struggle to handle long motion sequences as a single input due to prohibitively high computational cost; 
(2) Breaking down the generation of long motion sequences into shorter segments can result in inconsistent transitions and requires interpolation or inpainting, which lacks entire sequence modeling. 
To solve these challenges, 
we propose \textbf{InfiniMotion}, a method that generates continuous motion sequences of arbitrary length within an autoregressive framework. We highlight its groundbreaking capability by generating a continuous \textbf{1-hour} human motion with around \textbf{80,000} frames.
Specifically, we introduce the Motion Memory Transformer with Bidirectional Mamba Memory, enhancing the transformer's memory to process long motion sequences effectively without overwhelming computational resources. 
Notably, our method achieves over \textbf{30\%} improvement in FID and \textbf{6} times longer demonstration compared to previous state-of-the-art methods, showcasing significant advancements in long motion generation.

\end{abstract}

\section{Introduction}

Creating 3D human movements based on textual descriptions, known as \textit{text-to-motion} generation \cite{guo2022generating}, has sparked considerable interest due to its various applications in fields including computer games, virtual and augmented reality, and film production. Recent advancements \cite{guo2022generating,tevet2022human,chen2023executing,zhang2024motion,guo2023momask} in text-to-motion generation using diffusion \cite{ho2020denoising, song2020denoising, dhariwal2021diffusion, nichol2021glide, rombach2022high} and autoregressive models \cite{vaswani2017attention, radford2018improving} have demonstrated promising results. In real-world scenarios, there are various situations that demand long motion generation lasting minutes or even hours, such as computer-generated imagery and animation. Additionally, online scenarios often require continuous and adaptable motion generation of flexible lengths, as seen in games and real-time robot manipulation. However, previous state-of-the-art methods typically focus on generating short motion clips, often lasting less than 10 seconds. This limitation significantly restricts the real-world applicability of text-guided human motion generation techniques, posing challenges for longer and more comprehensive motion sequences in practical scenarios. The main challenges in generating arbitrary long motion are twofold:

One significant challenge arises from the limitation of the current architectures of methods \cite{guo2022generating, tevet2022motionclip, tevet2022human, chen2023executing,guo2023momask}, which cannot process a long motion sequence as a whole.  VAE-based \cite{petrovich2022temos,guo2022generating} architectures compress a motion sequence into a lower-dimensional space and then project it back to the original motion space. The longer the input motion sequence, the more information is lost because the capacity of the latent space to hold information is limited \cite{kingma2019introduction,kingma2013auto}. 
Diffusion-based generative models \cite{zhang2024motiondiffuse,tevet2022human,chen2023executing} can only generate motions with a fixed number of frames, but in real-world scenarios, the required length of generated motion varies.
Transformer-based methods \cite{zhang2023t2m, guo2023momask} often involve challenges when processing the whole long motion sequence, as the quadratic complexity of the attention mechanism \cite{vaswani2017attention} significantly increases the computational load and the demand for resources.

Another significant challenge arises from the datasets they attempted on \cite{guo2022generating, plappert2016kit}, which consist solely of motion sequences within 200 frames. This limitation severely restricts the model’s ability to handle longer motion sequences, hindering its development in this area. Exploring datasets including longer sequences would be essential for overcoming this limitation and improving the model's performance in generating long motion sequences.

To address these challenges, current long motion generation methods incorporate diffusion models to perform interpolation and inpainting sampling, ensuring smooth transitions and continuity between adjacent motion segments. For instance, PriorMDM \cite{shafir2023human} uses a two-stage process to improve segment transitions by first creating a transition interval to align segments and then applying linear masking during denoising to refine the transitions. In addition, PCMDM \cite{yang2023synthesizing} generates a motion segment based on the previous one using past inpainting sampling and refines transitions with compositional transition sampling. However, these methods prioritize transitions between adjacent segments but neglect semantic coherence across the entire motion sequence. More importantly, they fail to address the application of online continuous motion generation, which is crucial for real-time and seamless motion generation in real-world application.

Meanwhile, autoregressive methods inherently possess the capability of processing consistent input streams and have already achieved notable success in natural language processing tasks \cite{vaswani2017attention, radford2018improving,radford2019language,brown2020language, achiam2023gpt}. Furthermore, memory-enhanced transformers \cite{martins2022former, dai2019transformer,gupta2020gmat,ainslie2020etc,burtsev2020memory,bulatov2022recurrent,bulatov2023scaling} have significantly improved the ability of transformers to model long contexts and retain long-term memory. Hence, developing a long motion generation method based on memory-enhanced transformers presents a promising direction. However, the memory within transformers experiences decay issues when handling extended input streams, a problem that becomes more pronounced as the input length increases.

Recent advancements in selective state space models, notably Mamba \cite{gu2023mamba}, have shown significant promise in long-sequence modeling, outperforming transformers \cite{yu2024mambaout}. Customizing Mamba for long sequence human motion generation has already yielded promising results \cite{zhang2024motion}. Therefore, utilizing the predominance of long sequence model of the Mamba model presents a promising direction to enhance the transformer's memory for long motion generation. To leverage these gaps, it is essential to develop a method that can seamlessly integrate both the overarching global semantics of the entire sequence and the specific local textual context of individual motion segments. The approach is desired to ensure accurate motion generation within each segment and facilitate smooth transitions between adjacent segments, without requiring additional interpolation processes between these segments. 

In response to these challenges, we propose the following three key contributions.

\begin{itemize}
    \item We propose a novel autoregressive motion generation method, named \textbf{InfiniMotion}, which is capable of generating continuous motion sequences of arbitrary length. To showcase the groundbreaking capabilities of our approach, we present a demonstration featuring a \textbf{1-hour} human motion sequence with around \textbf{80,000} frames generated by our proposed method. This sequence corresponds to over \textbf{400} segment-level motion queries.
    \item To tackle the long motion generation, we introduced the Motion Memory Transformer, specifically designed for handling long-context motion generation by leveraging additional memory capacity to the motion transformers. To further enhance long-term memory and maintain coherence throughout the entire motion sequence, we incorporated the Bidirectional Mamba Memory block to improve the memory capacity of the Transformer architecture. This mechanism effectively addresses long-term memory forgetfulness, ensuring both the overarching global semantics of the entire sequence and smooth transitions between adjacent segments.
    \item We conducted a thorough evaluation of the extensive human motion dataset BABEL. Our approach achieved an over \textbf{30\%} improvement in the FID and demonstrated a motion sequence that is \textbf{6} times longer than those produced by previous state-of-the-art methods \cite{yang2023synthesizing,shafir2023human}.  This advancement highlights a significant improvement in our method for maintaining consistency over long motion sequences and extending temporal length. These promising results represent a significant improvement in human motion generation, bringing it closer to real-world applications.
\end{itemize}

\section{Related Work}

\paragraph{Text-to-Motion Generation.}

Text-to-motion generation aims to generate motions that align with provided textual instructions. Advances in multi-modal pretraining \cite{radford2021learning} and generative models \cite{ho2020denoising, rombach2022high} have greatly advanced text-to-motion generation, garnering considerable interest. JL2P \cite{ahuja2019language2pose} utilizes RNN-based autoencoders to grasp a joint representation of language and posture, enforcing a direct one-to-one relationship between text and motion. MotionCLIP \cite{tevet2022motionclip} employs transformer-based \cite{vaswani2017attention} autoencoders to rebuild motion sequences, ensuring they correspond accurately to the accompanying text descriptions within the CLIP \cite{radford2018improving} space. This alignment effectively merges semantic knowledge from CLIP into the human motion domain. TEMOS \cite{petrovich2022temos} and T2M \cite{guo2022generating} integrate a Transformer-based VAE with a text encoder to generate distribution parameters operating within the VAE latent space. AttT2M \cite{zhong2023attt2m} and TM2D \cite{gong2023tm2d} enhance learning by incorporating a body-part spatio-temporal encoder into VQ-VAE, thereby enriching the discrete latent space with greater expressiveness. MotionDiffuse \cite{zhang2024motiondiffuse} presents the first framework for generating motion based on text input using diffusion models. It highlights key features such as probabilistic mapping, realistic synthesis, and multi-level manipulation. Meanwhile, MDM \cite{tevet2022human} employs a Transformer-based diffusion model specifically designed for human motion, aiming to predict samples rather than noise at each step. On the other hand, MLD \cite{chen2023executing} conducts diffusion in latent motion space instead of establishing connections between raw motion sequences and conditional inputs using a diffusion model. T2M-GPT \cite{zhang2023t2m} combines VQ-VAE and transformers for human motion generation from text, outperforming recent diffusion-based methods. Motion Mamba \cite{zhang2024motion} introduced an efficient architecture for long-sequence human motion generation by incorporating state space models (SSMs) \cite{gu2023mamba} and diffusion. MMM \cite{pinyoanuntapong2023mmm} introduced a mask motion model which achieved high-quality and real-time motion generation with editing features. MoMask \cite{guo2023momask} introduces a masked transformer architecture with residual transformers for text-driven human motion generation. This approach is applicable to related tasks without requiring additional fine-tuning and has been widely adopted in animated avatar generation \cite{zhang2024motionavatar}.

\paragraph{Long-sequence Motion Generation.}

Previous long motion generation methods typically employ recurrent architectures to process motion segments recursively or generate separate motion segments and interpolate between them using diffusion models. Long motion generation based on RNNs often leads to the problem of collapsing into repetitive or static poses. The acRNN \cite{li2017auto} addresses this challenge by training the model with its own generated frames for prefix completion tasks, although it still faces limitations with the relatively short sequences of available data. Recent works propose overcoming data limitations by generating short sequences in an auto-regressive manner, with each sequence conditioned on both a textual prompt and the ending of the previous sequence. Transitions were learned either based on a smoothness prior, such as TEACH \cite{athanasiou2022teach}, or directly from the data, as seen in MarioNet \cite{wang2022neural}. The Perpetual Motion \cite{zhang2020perpetual} claims its capability to generate potentially infinite motion sequences utilizing a two-stream variational RNN model. Nonetheless, this approach is limited to motion prediction based on historical motion data and does not support text-to-motion generation. The Story-to-Motion \cite{qing2023story} introduces a motion retrieval method capable of synthesizing infinite and controllable human motions based on a storyline. However, it is limited to extracting existing motion segments from a database and cannot achieve text-to-motion generation. PCMDM \cite{yang2023synthesizing} advanced long sequence text-to-motion generation, overcoming limitations of existing methods by proposing a novel approach utilizing past-conditioned diffusion model and coherent sampling techniques. PriorMDM \cite{shafir2023human} introduced a novel sequential composition method that generates extended motions by assembling sequences of prompted intervals and their transitions, utilizing a prior trained solely on short clips.

\section{Methodology}
\label{sec:method}

\begin{figure}[t]
    \centering
    \includegraphics[width=1\linewidth]{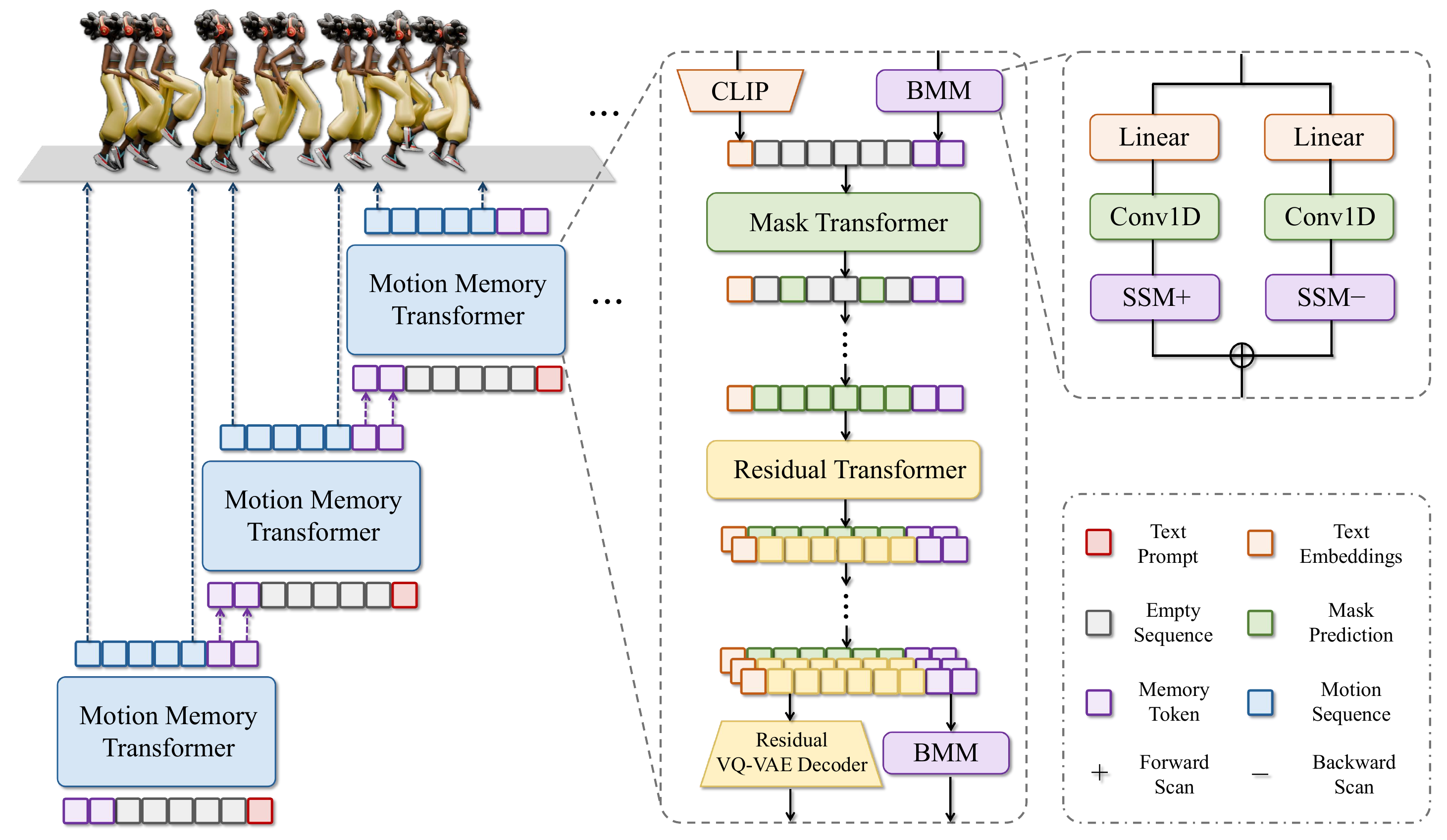} 
    \caption{This diagram illustrates the main architecture of our proposed method. The method processes a stream of motion segments in an autoregressive manner within a recurrent memory architecture. The Motion Memory Transformer (MMT) enhances each motion segment with a specialized memory token \texttt{[mem]}, which facilitates both long-term semantic coherence and smooth transitions between adjacent motion segments based on user text queries. Within the MMT, we leverage the robust long-term memory capabilities of Mamba \cite{gu2023mamba}, and we have customized a Bidirectional Mamba Memory (BMM) block to further enhance the memory within the transformer. This customization ensures long-term coherence that corresponds to the overall semantics of the entire motion sequence.}
    \label{fig:main}
\end{figure}

\subsection{Problem Formulation}

We tackle the challenge of text-conditioned long-sequence human motion generation. Specifically, our approach takes a set of text queries \( Q = \{q_1, q_2, q_3, \ldots q_p \} \) as input and generates the corresponding long-sequence 3D human motions, where \( p \) denotes the number of motion segments. The objective of this task is to generate a high-quality, coherent long-sequence motion \( \hat{Y} = \{\hat{y}_1, \hat{y}_2, \hat{y}_3, \ldots, \hat{y}_p\} \) corresponding to the text queries \( Q \).

\subsection{Preliminaries}

\paragraph{Residual VQ-VAE.} The backbone of the motion generation model is adapted from MoMask \cite{guo2023momask}. Given a training motion sequence \( Y = \{y_1, y_2, y_3, \ldots, y_p\} \), where each motion segment \( y_i \in \mathbb{R}^{N \times D} \) for \( i \in [1, p] \), contains \( N \) frames and has a motion dimension of \( D \). For each motion segment, the residual VQ-VAE encoder \(E_{rvq}\) compresses a motion segment into a latent vector \(\mathbf{\tilde{b}} \in \mathbb{R}^{n \times d}\), with a downsampling ratio of \(n/N\) and a latent dimension \(d\).
\begin{equation}
    \mathbf{\tilde{b}} = E_{rvq}(y_i).
\end{equation}
Then, the vector $\mathbf{\tilde{b}}$ is replaced with its nearest code entry in a codebook $\mathcal{C} = \{{\mathbf{c}_k}\}^K_{k=1} \subset \mathbb{R}^d$ using $V + 1$ quantization layers. This process is commonly referred to as residual quantization $RQ(\cdot)$. Formally, this is defined as 
\begin{equation}
    RQ(\mathbf{\tilde{b}}) = [\mathbf{b}^v]_{v=0}^V, 
\end{equation}
where $\mathbf{b}^v \in \mathbb{R}^{n \times d}$ represents the code sequence at the $v$-th quantization layer. Hence, the final approximation of the latent vector $\mathbf{\tilde{b}}$ is the sum of all quantized sequences $\mathbf{b}^v$. 
\begin{equation}
    \mathbf{\tilde{b}} = \sum^V_{v=0}\mathbf{b}^v.
\end{equation}
Furthermore, each motion segment can be represented as $V + 1$ discrete motion token sequences, denoted as 
\begin{equation}
    T = [t^v]^V_{v=0},
\end{equation}
where each token sequence $t^v \in \{1,...,|\mathcal{C}^v|^n\}$ represents the ordered codebook-indices of quantized embedding vectors $\mathbf{b}^v$. Later, the latent vector $\mathbf{b}$ will be projected back to the motion space for motion reconstruction using the residual VQ-VAE decoder $D_{rvq}$.
\begin{equation}
    \hat{y_i} = D_{rvq}(\mathbf{b}).
\end{equation}

\paragraph{Mask Transformer.} Within latent space, each text embedding $e_i$ is extracted from prompt text query $q_i$ for $i \in [1,p]$ using CLIP \cite{radford2021learning}, and the mask transformer (MT) draws inspiration from the bidirectional mask mechanism found in BERT \cite{devlin2018bert}. Initially, the mask transformer randomly replaced a varying portion of sequence elements $t^0 \in \mathbb{R}^n$ with a special \texttt{[MASK]} token on temporal dimension $k$, aiming to predict these masked tokens based on text embedding $e_i$ and sequence after masking $\tilde{t}^0$. The mask transformer is optimized to minimize the negative log-likelihood of target predictions:
\begin{equation}
    \mathcal{L}_{mask} = \sum_{\tilde{t}^0_k = \texttt{[MASK]}} - \log \text{MT}(t^0_k|\tilde{t}^0,e_i).
\end{equation}

\paragraph{Residual Transformer.} The residual transformer (RT) is trained to model tokens from $V$ other residual quantization layers. During training, a quantizer layer \( j \in [1,V] \) is randomly selected. Subsequently, all tokens in the preceding layers \( t^{0:j-1} \) are embedded and aggregated as the token embedding input. Certainly: By utilizing token embedding, text embedding, and the RQ layer indicator $j$ as input, the residual transformer is trained to predict the tokens of the $j$-th layer simultaneously. Hence, the optimization objective can be represented as:
\begin{equation}
    \mathcal{L}_{res} = \sum_{j=1}^V \sum_{l=1}^n - \log \text{RT}(t_l^j|t_l^{1:j-1},e_i,j).
\end{equation}

\subsection{Motion Memory Transformer}

Autoregressive methods tailored with memory-enhanced transformers inherently possess the capability to process consistent input streams. They notably enhance the ability of transformers to model long contexts and retain long-term memory. Inspired by Recurrent Memory Transformer \cite{bulatov2022recurrent,bulatov2023scaling}, the Motion Memory Transformer (MMT) comprises a series of masked transformers and residual transformers, shown in Figure \ref{fig:main}. The token sequence \( t^v \) of each motion segment $y_i$ is first concatenated with the text embedding \( e_i \) to serve as the input $M_i$ to the transformer, where $\circ$ denotes concatenation.

\begin{equation}
    M_i = [e_i \circ t^v], \text{for } i \in [1, p].
\end{equation}

The input $M_i$ is augmented with special \texttt{[mem]} tokens, which are processed in conjunction with the sequence of tokens using standard techniques. Since the mask transformer is an encoder-only model that utilizes bidirectional masking similar to BERT, rather than a decoder-only model with causal attention masking, we only need to add a single memory component to the input.

\begin{equation}
    \tilde{M_i} = [M_i \circ M^{mem}].
\end{equation}

Subsequently, the Motion Memory Transformer takes \(\tilde{M_i}\) as input and produces \(\hat{M_i}\), which comprises logits \(\bar{M_i}\) and the memory token \(M^{mem}\).

\begin{equation}
    \hat{M_i} = \text{MMT}(\tilde{M_i}), \hat{M_i} \coloneqq [\bar{M_i} \circ M^{mem}].
\end{equation}

After processing by transformers, the \texttt{[mem]} tokens are concatenated with the next motion segment $M_{i+1}$. This concatenated sequence is then subjected to the same recurrent processing steps until the entire motion sequence has been processed.

\begin{equation}
    \tilde{M}_{i+1} = [M_{i+1} \circ M^{mem}].
\end{equation}

\subsection{Bidirectional Mamba Memory}

Despite memory-enhanced transformers are able to preserve long-term dependencies across multiple input segments, they often experience memory decay and forgetfulness as the number of input segments increases. However, selective state space models, particularly Mamba \cite{gu2023mamba}, have demonstrated promising advantages in long-sequence modeling compared to transformers \cite{yu2024mambaout}. Hence, utilizing the robust memory capabilities of Mamba offers a promising approach for long motion generation. Inspired by Motion Mamba \cite{zhang2024motion}, we designed the Bidirectional Mamba Memory (BMM) block within the Motion Memory Transformer to enhance memory retention. This enhancement supports both the transitions between adjacent motion segments and the long-term coherence of entire motion sequences. 

In the Motion Memory Transformer, the \texttt{[mem]} token undergoes a linear projection following normalization, as illustrated on the right side of Figure \ref{fig:main}. After the memory token is subjected to a 1-D convolution, it processed by SSMs scanning in both forward and backward directions. This technique enhances the receptive field and removes the bias associated with the order of input memories, as the memory token itself does not have explicit temporal order. The outputs of SSMs undergo subsequent gating and summation processes, culminating in the production of the BMM block as an enhanced memory token.

\section{Experiment}
\label{sec:exp}

\subsection{Dataset and Evaluation Metrics}

\paragraph{BABEL Dataset.} In conventional text-to-motion datasets \cite{guo2022generating, plappert2016kit}, individual motion sequences are typically linked to singular textual descriptions. These motion clips tend to be brief, frequently containing fewer than 200 frames. Such brevity renders them unsuitable for tasks requiring the generation of long motion sequences. The BABEL dataset \cite{punnakkal2021babel}, derived from AMASS \cite{mahmood2019amass}, offers finely detailed annotations for extended motion sequences, wherein each segment correlates with a distinct textual annotation. This dataset encompasses a total of 10,881 motion sequences, comprising 65,926 segments along with their respective textual labels. There are two primary methods for processing the BABEL dataset. PriorMDM \cite{shafir2023human} followed TEACH \cite{athanasiou2022teach}, which curated a subset of BABEL by excluding poses like `a-pose' or `t-pose' and combined transitions with the second segment. While PCMDM \cite{yang2023synthesizing} reconstructed BABEL into motion segments pairs, derived from adjacent pairs of segments within long sequences. For instance, given a motion sequence in the BABEL dataset such as [`walk', `sit down', `wave right hand'], the PCMDM processed it into two motion segment pairs: [`walk', `sit down'] and [`sit down', `wave right hand'].

\paragraph{Evaluation Metrics.} The quantitative evaluation of text-to-motion generation, as introduced by T2M \cite{guo2022generating}, is adopted throughout our experiments. This includes: (1) \textit{Frechet Inception Distance} (FID), which assesses the overall motion quality by measuring the distributional difference between the high-level features of generated motions and real motions; (2) \textit{R-precision} and (3) \textit{MultiModal Distance}, which evaluate the semantic alignment between the input text and generated motions; and (4) \textit{Diversity}, which calculates the variance in features extracted from the motions.

\subsection{Comparative Studies}

To evaluate the performance of our model on long-sequence motion generation, we first pre-trained the model on the HumanML3D dataset \cite{guo2022generating} at the motion segment level. Subsequently, the model was trained and evaluated on two distinct processed versions of the BABEL dataset \cite{punnakkal2021babel}. We further leveraged Flash Attention \cite{dao2022flashattention, dao2023flashattention} to enhance the speed of training and inference in long motion generation tasks. The results presented in Tables \ref{tab:priormdm_babel} and \ref{tab:pcmdm_babel} demonstrate that our method significantly outperforms the previous text-to-motion generation approaches, which are specifically designed for long-sequence motion generation. 
All experiments were conducted using a batch size of 256 for RVQ-VAE with 6 quantize layers, and a batch size of 64 for MMT, with a memory token length of 128. These experiments were performed on a single Intel Xeon Platinum 8360Y CPU 2.40GHz, paired with a single NVIDIA A100 40G GPU and 32GB of RAM.

\begin{table}[htbp]
\centering
\caption{This table illustrates a comparison between our method and previous long motion generation techniques on the PriorMDM-modified BABEL dataset \cite{punnakkal2021babel} to ensure fairness. The results indicate that our method outperforms these techniques, demonstrating superior performance. The right arrow $\rightarrow$ means that the closer to the real motion, the better. \textbf{Bold} and \underline{underline} indicate the best and second best result.}
\resizebox{\columnwidth}{!}{%
\begin{tabular}{lcccc}
  \toprule
Method & R-precision $\uparrow$ & FID $\downarrow$ & Diversity $\rightarrow$ & MultiModal-Dist $\downarrow$\\
\midrule
Ground Truth  & ${0.62}$ & $0.4 \cdot 10^{-3}$ & ${8.51}$ & ${3.57}$\\
\midrule
TEACH \cite{athanasiou2022teach} & $0.46$ & $1.12$ & $8.28$ & $7.14$ \\
PriorMDM \cite{shafir2023human} & $ 0.43$ & $1.04$ & $8.14$& $7.39$\\
PriorMDM w/ Trans. Emb \cite{shafir2023human} 
& $0.48$ & $0.79$ & $8.16$ & $6.97$\\
PriorMDM w/ Trans. Emb \& geo losses \cite{shafir2023human} 
& $0.45$ & $0.91$ & $8.16$& $7.09$\\
Motion Mamba \cite{zhang2024motion} & $\underline{0.49}$ & $\underline{0.76}$ & $\mathbf{8.39}$ & $\underline{4.97}$  \\

\midrule

\textbf{InfiniMotion (Ours)} & $\mathbf{0.51}$ & $\mathbf{0.58}$ & $\underline{8.67}$ & $\mathbf{4.89}$ \\

\bottomrule
\end{tabular}}
\label{tab:priormdm_babel}
\end{table}

\begin{table}[tbp]
\centering
\caption{The table demonstrates the comparison of our method with previous long motion generation methods on the PCMDM-modified BABEL dataset \cite{punnakkal2021babel}, ensuring a fair evaluation. The results demonstrate the superior performance of our method over these approaches. The right arrow $\rightarrow$ means that the closer to the real motion, the better. \textbf{Bold} and \underline{underline} indicate the best and second best result.}
\resizebox{\columnwidth}{!}{%
\begin{tabular}{lcccc}
\toprule
Method & FID $\downarrow$ & R-precision (Top 3) $\uparrow$ & MultiModal-Dist $\downarrow$ & Diversity $\rightarrow$ \\ 
\midrule
Ground Truth & $0.009^{\pm0.001}$ &$0.773^{\pm0.002}$ & $21.860^{\pm0.006}$ & $15.034^{\pm0.078}$ \\
\midrule
TEACH-Independent \cite{athanasiou2022teach} & $12.256^{\pm.0.284}$ & $\underline{0.816}^{\pm.0.004}$ & ${21.874}^{\pm.0.115}$ & ${13.905}^{\pm.0.066}$ \\
TEACH-Joint \cite{athanasiou2022teach} & $13.084^{\pm.0.284}$ & ${0.783}^{\pm.0.008}$ & ${22.218}^{\pm.0.103}$ & ${13.624}^{\pm.0.094}$ \\
TEACH \cite{athanasiou2022teach} & $7.312^{\pm.0.019}$ &$\mathbf{0.864}^{\pm.0.005}$ & $21.017^{\pm.0.078}$ & ${14.248}^{\pm.0.070}$ \\
MDM \cite{tevet2022human} & ${7.476}^{\pm0.232}$ &$0.770^{\pm0.0.008}$ & $22.020^{\pm0.098}$ & $13.876^{\pm.0.071}$ \\
MDM w/ Inpainting Sampling \cite{tevet2022human} & $7.411^{\pm.0.242}$ &${0.772}^{\pm.0.005}$ & ${22.038}^{\pm.0.109}$ & ${13.788}^{\pm.0.083}$ \\
MDM w/ Compositional Sampling \cite{tevet2022human} & $7.057^{\pm.0.232}$ &${0.782}^{\pm.0.006}$ & ${21.868}^{\pm.0.110}$ & ${14.112}^{\pm.0.087}$ \\
PCMDM \cite{yang2023synthesizing} & ${5.396}^{\pm.0.187}$ &${0.775}^{\pm.0.010}$ & ${21.653}^{\pm.0.120}$ & ${14.534}^{\pm.0.063}$ \\
PCMDM w/ Inpainting Sampling \cite{yang2023synthesizing} & $5.431^{\pm.0.176}$ &${0.778}^{\pm.0.008}$ & ${21.646}^{\pm.0.126}$ & ${14.501}^{\pm.0.084}$ \\
PCMDM w/ Compostional Sampling \cite{yang2023synthesizing} & $5.242^{\pm.0.131}$ &${0.799}^{\pm.0.007}$ & ${21.412}^{\pm.0.087}$ & $14.652^{\pm.0.068}$ \\
Motion Mamba \cite{zhang2024motion} & $\underline{3.637}^{\pm.0.230}$ & ${0.797}^{\pm.0.003}$ & $\underline{6.301}^{\pm.0.116}$ & $\underline{14.685}^{\pm.0.050}$ \\
\midrule
\textbf{InfiniMotion (Ours)} & $\mathbf{2.909}^{\pm.0.165}$ & ${0.801}^{\pm.0.008}$ & $\mathbf{5.042}^{\pm.0.073}$ & $\mathbf{14.761}^{\pm.0.039}$ \\
\bottomrule
\end{tabular}}
\label{tab:pcmdm_babel}
\vspace{-0.2cm}
\end{table}

\subsection{Ablation Studies}

To further investigate the impact of varying the length of the memory token \texttt{[mem]} on the performance of long motion generation, we conducted ablation studies using different lengths of memory tokens in the Motion Memory Transformer on the PriorMDM-modified BABEL dataset. As shown in Table \ref{tab:mem}, the results indicate that the optimal performance is achieved with a memory token length of 128.

\begin{table}[tbp]
\centering
\caption{The following table presents an ablation study examining the influence of varying the length of the memory token \texttt{[mem]} on the efficacy of long motion generation. The investigation was carried out utilizing PriorMDM-customized BABEL. The findings indicate that the most favorable performance is attained when employing a memory token length of 128. The right arrow $\rightarrow$ means that the closer to the real motion, the better. \textbf{Bold} and \underline{underline} indicate the best and second best result. }
\resizebox{0.9\columnwidth}{!}{%
\begin{tabular}{lcccc}
  \toprule
Method & R-precision $\uparrow$ & FID $\downarrow$ & Diversity $\rightarrow$ & MultiModal-Dist $\downarrow$\\
\midrule
Ground Truth  & ${0.62}$ & $0.4 \cdot 10^{-3}$ & ${8.51}$ & ${3.57}$\\
\midrule
InfiniMotion w/o \textit{mem} & $0.44$ & $0.74$ & $8.33$ & $5.31$ \\
InfiniMotion (\textit{mem} = 32) & $0.47$ & $0.65$ & $8.71$ & $5.52$ \\
InfiniMotion (\textit{mem} = 64) & $0.48$ & $0.63$ & $\underline{8.69}$ & $5.13$ \\
InfiniMotion (\textit{mem} = 196) & $\underline{0.50}$ & $\mathbf{0.57}$ & $8.71$ & $\underline{4.94}$ \\
InfiniMotion (\textit{mem} = 256) & $\mathbf{0.51}$ & $\underline{0.58}$ & $\mathbf{8.67}$ & $5.01$ \\
\midrule
\textbf{InfiniMotion (\textit{mem} = 128)} & $\mathbf{0.51}$ & $\underline{0.58}$ & $\mathbf{8.67}$ & $\mathbf{4.89}$ \\
\bottomrule
\end{tabular}}
\label{tab:mem}
\end{table}

To investigate the impact of the Bidirectional Mamba Memory (BMM) block on both the overall performance of long motion generation and its relation to the number of input segments, we conducted a series of thorough experiments as detailed below. We conducted three distinct sets of experiments along with the overall evaluation using different subsets of the PriorMDM-modified BABEL dataset. The dataset subsets were divided based on the number of segments ($N$), specifically: $N < 5$, $5 \leq N < 10$, and $N \geq 10$. The results presented in Table \ref{tab:bmm} demonstrate two key findings: (1) The inclusion of the BMM block significantly enhances long motion generation performance across all quantitative metrics compared to the configuration without the BMM block. (2) The configuration without the BMM block exhibits a slight decrease in performance as the number of input segments decreases. In contrast, the configuration with the BMM block effectively addresses this issue, maintaining superior performance across varying input lengths. This stability underscores the BMM block's substantial positive impact on enhancing memory within the transformer model.

\begin{table}[htbp]
\centering
\caption{The table demonstrates the impact of the Bidirectional Mamba Memory (BMM) block on the overall performance of long motion generation and its relation to the number of input segments. The results show that the BMM block effectively enhances the transformer's memory, maintaining superior performance across varying input lengths. The right arrow $\rightarrow$ means that the closer to the real motion, the better. \textbf{Bold} indicates the best result. The \textcolor{Green}{\(\blacktriangle\)} denotes an improvement compared to the aforementioned method.}
\resizebox{0.9\columnwidth}{!}{%
\begin{tabular}{lcccc}
  \toprule
Method & R-precision $\uparrow$ & FID $\downarrow$ & Diversity $\rightarrow$ & MultiModal-Dist $\downarrow$\\
\midrule
Ground Truth  & ${0.62}$ & $0.4 \cdot 10^{-3}$ & ${8.51}$ & ${3.57}$\\
\midrule
InfiniMotion w/o BMM ($N < 5$) & $0.49$ & $0.74$ & $8.31$ & $5.31$ \\
\textbf{InfiniMotion w/ BMM ($N < 5$)} & $\mathbf{0.51} \textcolor{Green}{\raisebox{-0.8ex}{\textsuperscript{$\blacktriangle \mathbf{0.02}$}}}$ & $\mathbf{0.59} \textcolor{Green}{\raisebox{-0.8ex}{\textsuperscript{$\blacktriangle \mathbf{0.15}$}}}$ & $\mathbf{8.68} \textcolor{Green}{\raisebox{-0.8ex}{\textsuperscript{$\blacktriangle \mathbf{0.03}$}}}$ & $\mathbf{4.83} \textcolor{Green}{\raisebox{-0.8ex}{\textsuperscript{$\blacktriangle \mathbf{0.48}$}}}$ \\
\midrule
InfiniMotion w/o BMM ($5 \leq N < 10$) & $0.47$ & $0.76$ & $8.25$ & $5.84$ \\
\textbf{InfiniMotion w/ BMM ($5 \leq N < 10$)} & $\mathbf{0.50} \textcolor{Green}{\raisebox{-0.8ex}{\textsuperscript{$\blacktriangle \mathbf{0.03}$}}}$ & $\mathbf{0.57} \textcolor{Green}{\raisebox{-0.8ex}{\textsuperscript{$\blacktriangle \mathbf{0.19}$}}}$ & $\mathbf{8.66} \textcolor{Green}{\raisebox{-0.8ex}{\textsuperscript{$\blacktriangle \mathbf{0.11}$}}}$ & $\mathbf{4.88} \textcolor{Green}{\raisebox{-0.8ex}{\textsuperscript{$\blacktriangle \mathbf{0.96}$}}}$ \\
\midrule
InfiniMotion w/o BMM ($N \geq 10$) & $0.45$ & $0.77$ & $8.27$ & $5.75$ \\
\textbf{InfiniMotion w/ BMM ($N \geq 10$)} & $\mathbf{0.52} \textcolor{Green}{\raisebox{-0.8ex}{\textsuperscript{$\blacktriangle \mathbf{0.07}$}}}$ & $\mathbf{0.57} \textcolor{Green}{\raisebox{-0.8ex}{\textsuperscript{$\blacktriangle \mathbf{0.20}$}}}$ & $\mathbf{8.69} \textcolor{Green}{\raisebox{-0.8ex}{\textsuperscript{$\blacktriangle \mathbf{0.06}$}}}$ & $\mathbf{4.94} \textcolor{Green}{\raisebox{-0.8ex}{\textsuperscript{$\blacktriangle \mathbf{0.81}$}}}$ \\
\midrule\midrule
InfiniMotion w/o BMM & $0.46$ & $0.76$ & $8.27$ & $5.69$ \\
\textbf{InfiniMotion w/ BMM} & $\mathbf{0.51} \textcolor{Green}{\raisebox{-0.8ex}{\textsuperscript{$\blacktriangle \mathbf{0.05}$}}}$ & $\mathbf{0.58} \textcolor{Green}{\raisebox{-0.8ex}{\textsuperscript{$\blacktriangle \mathbf{0.18}$}}}$ & $\mathbf{8.67} \textcolor{Green}{\raisebox{-0.8ex}{\textsuperscript{$\blacktriangle \mathbf{0.08}$}}}$ & $\mathbf{4.89} \textcolor{Green}{\raisebox{-0.8ex}{\textsuperscript{$\blacktriangle \mathbf{0.80}$}}}$ \\
\bottomrule
\end{tabular}}
\label{tab:bmm}
\end{table}

\subsection{Qualitative Evaluation}

\begin{figure}[H]
    \centering
    \includegraphics[width=\linewidth]{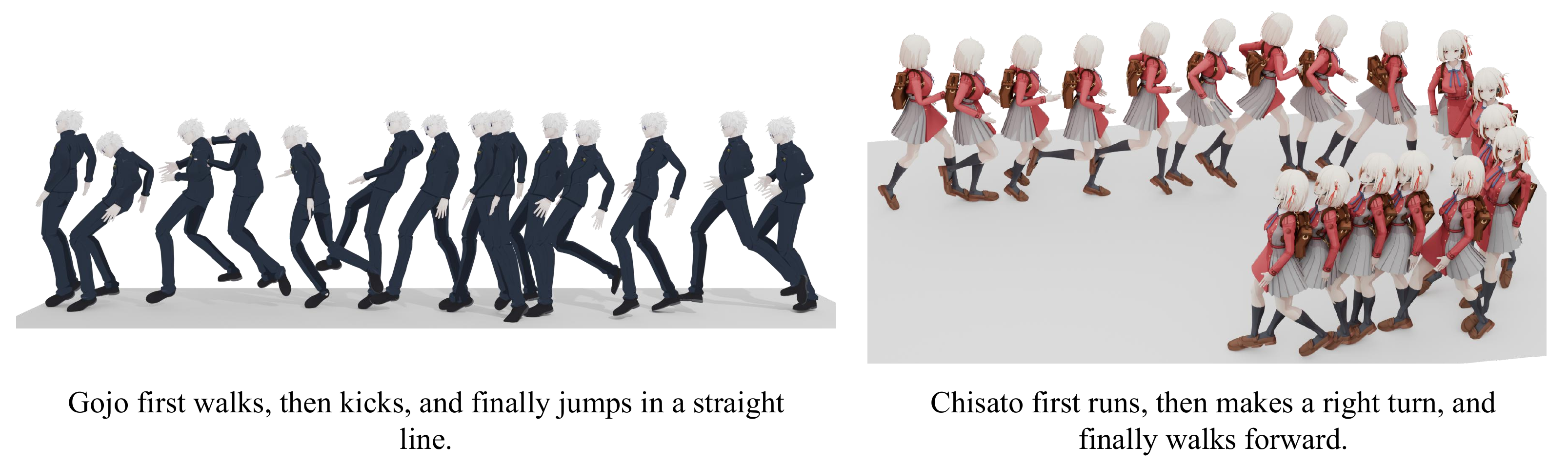} 
    \caption{The diagram presents additional examples of long motion sequences generated by our proposed method. These examples highlight the method's ability to produce smooth transitions between motion segments, resulting in high-quality and diverse motion outputs.}
    \label{fig:main_demo-3}
\end{figure}

We conducted a comprehensive quantitative evaluation of our method, encompassing two key aspects. First, we visualized long motion generation under various text conditions and different numbers of query segments. Figures \ref{fig:main_demo} and \ref{fig:main_demo-3} illustrate examples that demonstrate the superior performance of our method in generating high-quality, complex, and long-sequence motions. These results highlight the promising potential of our approach for real-world applications. Secondly, we conducted a comprehensive user study involving 50 participants to evaluate our long motion generation. The study was carried out using a Google Forms survey, as detailed in Appendix Section \ref{sec:user}. The results indicate overall positive feedback regarding both motion accuracy and user engagement, demonstrating promising advancements in long-term motion generation.

\section{Limitation and Failure Cases}
\label{sec:limit}

However, despite achieving convincing results, text-to-motion generation faces several limitations that hinder its real-world application. The two main limitations are: (1) The text condition is very general and sometimes vague, making it impractical and nearly impossible to achieve joint-level or more precise control of human motions. (2) The motion representation used in text-to-motion generation is the relative pose in each frame, typically represented as Euler rotation angles in 6D relative to the root joint (usually the hip or pelvis). Notable examples include SMPL \cite{SMPL:2015} and HumanML3D \cite{guo2022generating}. This representation poses challenges for users who need precise physical quantities such as velocity, distance, and rotation angle, leading to related failure cases.

\section{Discussion and Conclusion}
In this paper, we proposed InfiniMotion, an innovative autoregressive method for generating continuous motion sequences of arbitrary length. We demonstrated its capability by generating a 1-hour human motion sequence with approximately 80,000 frames, equivalent to over 400 segment-level motion queries. Our Motion Memory Transformer, enhanced by the Bidirectional Mamba Memory block, effectively handles long-context motion generation and addresses long-term memory forgetfulness, ensuring coherent and smooth transitions. Evaluations on the BABEL dataset showed over a 30\% improvement in FID and produced motion sequences 6 times longer than those by previous state-of-the-art methods, significantly advancing human motion generation towards real-world applications.

\textbf{Broader Impacts.}
Our work stands as a groundbreaking effort in the realm of customizing memory-enhanced transformer architecture for long motion generation. By incorporating state-space models (SSMs) to augment the transformer's memory capabilities, we have achieved promising results that demonstrate high-quality long-motion generation. This advancement holds substantial potential for transformative applications in video production, computer games, and robotic manipulation, marking a significant step forward in these fields.

\clearpage


\newpage
\appendix
\section*{Appendix}
\section{User Study}
\label{sec:user}
\begin{figure}[ht]
    \centering
    \includegraphics[width=\linewidth]{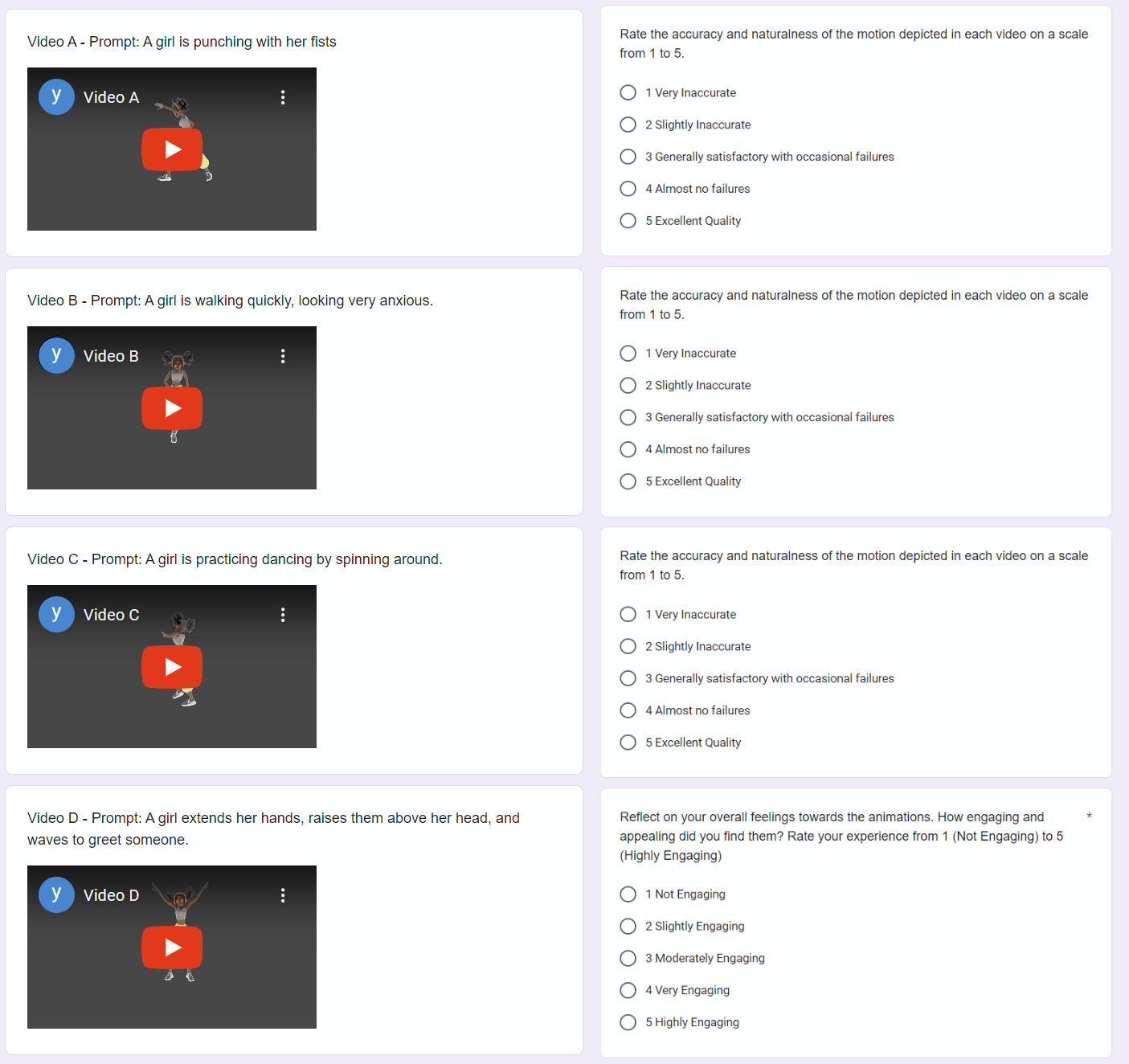}
    \vspace{20pt} %
    \caption{This figure displays the User Interface (UI) used in our User Study, showcasing four videos (Video A to D) each with distinct motion animations from the same model. Participants evaluate these animations on aspects such as motion accuracy, and overall user experience. They rate each aspect from 1 (low) to 5 (high) to assess how the animations mirror real-world movements and their engagement level. This evaluation aims to determine the realism and engagement effectiveness of each motion.
}
    \label{fig:user}
\end{figure}

This study offers a thorough evaluation of our motion generation, leveraging both qualitative and quantitative analyses. We assessed the real-world applicability of four motion videos produced by the Long Motion Generation platform, evaluated by 50 participants through a Google Forms survey as in Fig~\ref{fig:user}.

Participants were shown four videos labeled Video A, Video B, Video C, and Video D. Each video featured a distinct motion generated using Long Motion Generation with various input prompts. The participants evaluated these videos by answering a series of targeted questions designed to assess the motion’s accuracy, the mesh’s visual quality, the integration of motion and mesh, and their overall emotional response to the animations. The evaluation centered on several key aspects:

\begin{enumerate}
    \item \textbf{Motion Accuracy}: Participants rated the naturalness and accuracy of the motions on a scale from 1 ('Very Inaccurate') to 5 ('Excellent Quality'). The average score across the four videos was approximately \textbf{4.1}, indicating a high fidelity in motion portrayal.
    
    \item \textbf{User Engagement and Appeal}: Participants reflected on their feelings towards the animations, rating their overall engagement and appeal from 1 ('Not Engaging') to 5 ('Highly Engaging'). The average engagement score was \textbf{4.4}, suggesting that the animations were highly engaging and appealing to the audience.
\end{enumerate}

\subsection*{Results}

\begin{itemize}
    \item \textbf{91\%} felt that minor adjustments were necessary before deployment.
    \item \textbf{6\%} of participants believed the animations could be directly utilized in real-world applications without significant modifications.
    \item \textbf{3\%} of participants believed that our method needs further adjustments before real-world deployment.
\end{itemize}

These findings highlight the exceptional quality, impressive outcomes, and extensive applicability of the motions. Almost all participants rated the generated videos highly, with minimal criticisms regarding their quality. The results were particularly striking, with many participants showing enthusiasm for the widespread implementation of these animations.

This extensive user study demonstrates that our motions not only meet but exceed user expectations in terms of quality, realism, and engagement. They are well-suited for various practical applications. The insights gained from this study will inform further enhancements, ensuring our motion generation continues to lead in both technological and artistic innovation.

\end{document}